\documentclass[conference]{IEEEtran}
\IEEEoverridecommandlockouts
\usepackage{cite}
\usepackage[skip=0pt]{caption}
\usepackage{booktabs}
\usepackage{url}
\usepackage{amsmath,amssymb,amsfonts}
\usepackage{algorithmic}
\usepackage{graphicx}
\usepackage{hyperref}
\usepackage{textcomp}
\usepackage{xcolor}
\usepackage{tabularray}
\usepackage{academicons}
\usepackage{fontawesome}  % FontAwesome for icons
\usepackage{hyperref}
\def\BibTeX{{\rm B\kern-.05em{\sc i\kern-.025em b}\kern-.08em
    T\kern-.1667em\lower.7ex\hbox{E}\kern-.125emX}}

\begin{document}
	
	\title{Enhanced Drought Analysis in Bangladesh:  A Machine Learning Approach for Severity Classification Using Satellite Data}
	
	\author{
		\IEEEauthorblockN{Tonmoy Paul 
			\IEEEauthorblockA{
				Department of E.E.E.\\
				BRAC University\\
				Dhaka, Bangladesh\\
				tonmoypaul2003@gmail.com
			}
		}
		\and
		\IEEEauthorblockN{Mrittika Devi Mati 
			\IEEEauthorblockA{
				Department of C.S.\\
				BRAC University\\
				Dhaka, Bangladesh\\
				matimrittika2001@gmail.com
			}
		}
		\and
		\IEEEauthorblockN{Md. Mahmudul Islam 
			\IEEEauthorblockA{
				Department of E.E.E.\\
				BRAC University\\
				Dhaka, Bangladesh\\
				mahmudul.islam@bracu.ac.bd}
		}
	}

	\maketitle
	
	\begin{center}
		\textit{
			Disclaimer: This is the preprint version of a paper accepted at the 2024 27th International Conference on Computer and Information Technology (ICCIT), organized by IEEE Bangladesh Section. The final version will be published in IEEE Xplore.
		}
	\end{center}

\begin{abstract}
	Drought poses a pervasive environmental challenge in Bangladesh, impacting agriculture, socio-economic stability, and food security due to its unique geographic and anthropogenic vulnerabilities. Traditional drought indices, such as the Standardized Precipitation Index (SPI) and Palmer Drought Severity Index (PDSI), often overlook crucial factors like soil moisture and temperature, limiting their resolution. Moreover, current machine learning models applied to drought prediction have been underexplored in the context of Bangladesh, lacking a comprehensive integration of satellite data across multiple districts. To address these gaps, we propose a satellite data-driven machine learning framework to classify drought across 38 districts of Bangladesh. Using unsupervised algorithms like K-means and Bayesian Gaussian Mixture for clustering, followed by classification models such as KNN, Random Forest, Decision Tree, and Naive Bayes, the framework integrates weather data (humidity, soil moisture, temperature) from 2012-2024.  This approach successfully classifies drought severity into different levels. However, it shows significant variabilities in drought vulnerabilities across regions which highlights the aptitude of machine learning models in terms of identifying and predicting drought conditions. 
	
\end{abstract}


\section{Introduction}

\label{section:introduction}
Drought is an inherently catastrophic event induced by the climatological factors, landscape, climate, topography, and water demand of a specific region. Drought is more frequent in Bangladesh due to high temperatures and low rainfall. Bangladesh is known as one of the largest deltas in the world, extremely vulnerable to Natural Disasters due to its Strategic location, Low-lying, Flat terrain, Population density, Poverty, Lack of Institutional setup, etc \cite{foroumandi-2022}. When precipitation levels are consistently below average, it disrupts the natural water cycle, leading to diminished groundwater recharge. Soil moisture is a critical factor in mitigating drought, as it stores water that plants rely on during dry periods. Reduced soil moisture due to lack of precipitation or high temperatures can lead to agricultural drought \cite{berg-2018}. 

Machine Learning(ML) is nowadays a very popular method for analyzing complex datasets and solving problems. Drought analysis itself, a complex task, needs a huge amount of time and effort to interpret its characteristics. Author in \cite{belayneh-2013} showed how ML can be treated to forecast drought. Unsupervised ML models have been introduced in drought analysis several times as shown by the authors of \cite{Lalika2024} \cite{Sundararajan2021}.  

However, this study tries to inaugurate an innovative approach to drought severity classification, utilizing the machine learning technique to analyze satellite-driven weather dataset. This framework combines unsupervised clustering algorithms with supervised classification techniques to accurately predict drought severity. \\
We have organized this paper as follows: Section \ref{section:literature_review} reviews previous works and their limitations. Our proposed methodologies are explained in Section \ref{section:methodologies}, and the machine learning algorithms we used are detailed in Section \ref{section:ml_algorithms}. The analysis and discussion of the results are presented in Section \ref{section:result_discussion}. Finally, Section \ref{section:conclusion} provides the overall conclusion of this paper.

\section{Literature Review}
\label{section:literature_review}
The literature on drought analysis from the perspective of Bangladesh is comparatively infrequent. Indeed, most of the studies traditionally focus on flood risk assessments because of this country’s historical inclination to flooding. Recent studies and research projects have focused on drought forecasting, particularly on drought indices employing machine learning. 

The concepts, characteristics, complex nature of drought and the various environmental factors that influence drought; drought indicators are also identified and predicted by implementing Prediction Models and Adopted Technologies \cite{nandgude-2023}.
Several indices of drought such that; Standardized Precipitation Index (SPI) \cite{gorgij-2021},Palmer Drought Severity Index(PDSI), Standardized Precipitation-Evapotranspiration Index
(SPEI)\cite{prodhan-2022} and Vegetation Health Index (VHI) \cite{pham-2022}. 

\begin{table*}[htbp]
	\centering
	\caption{Advantages and Limitations to previously used drought index}
	\label{tab:advantages_disadvantages}
	\begin{tabular}{lllll} 
		\toprule
		\textbf{Index}         & \textbf{Focus}                                                             & \textbf{Calculation}                                                                                                                & \textbf{Advantages}                                                                                        & \textbf{Limitations}                                                                        \\ 
		\hline
		\textbf{SPI}  & \begin{tabular}[c]{@{}l@{}}Precipitation \\anomalies\end{tabular}          & \begin{tabular}[c]{@{}l@{}}Measures deviation \\from long-term averages\end{tabular}                                                & \begin{tabular}[c]{@{}l@{}}Simple to calculate, \\widely used\end{tabular}                                 & \begin{tabular}[c]{@{}l@{}}Does not consider other \\factors like temperature\end{tabular}  \\ 
		\hline
		\textbf{PDSI} & \begin{tabular}[c]{@{}l@{}}Moisture \\balance\end{tabular}                 & \begin{tabular}[c]{@{}l@{}}Considers precipitation, \\temperature, \\potential evapotranspiration, \\and soil moisture\end{tabular} & \begin{tabular}[c]{@{}l@{}}Accounts for multiple factors, \\provides comprehensive assessment\end{tabular} & \begin{tabular}[c]{@{}l@{}}Requires more data and \\complex calculations\end{tabular}       \\ 
		\hline
		\textbf{SPEI} & \begin{tabular}[c]{@{}l@{}}Moisture balance \\and temperature\end{tabular} & \begin{tabular}[c]{@{}l@{}}Combines precipitation and \\potential evapotranspiration\end{tabular}                                   & \begin{tabular}[c]{@{}l@{}}More sensitive to drought in \\regions with \\high temperatures\end{tabular}    & \begin{tabular}[c]{@{}l@{}}Requires more data and \\complex calculations\end{tabular}       \\ 
		\hline
		\textbf{VHI}  & \begin{tabular}[c]{@{}l@{}}Vegetation response \\to drought\end{tabular}   & \begin{tabular}[c]{@{}l@{}}Uses remote sensing data to \\assess vegetation health\end{tabular}                                      & \begin{tabular}[c]{@{}l@{}}Provides a direct measure of \\drought's impact\end{tabular}                    & \begin{tabular}[c]{@{}l@{}}Can be influenced by \\factors other than drought\end{tabular}   \\
		\bottomrule
	\end{tabular}
\end{table*}

The table \ref{tab:advantages_disadvantages} presents four commonly used drought indices: SPI, PDSI, SPEI, and VHI. Each index focuses on a different aspect of drought, brief overview of the calculation methods involved, advantages and limitations. For example, SPI is simple to calculate but does not consider temperature, while PDSI provides a comprehensive assessment but requires more data. The preference of index depends on the earmarked research question and the available data.

Moreover, types of drought Meteorological Drought\cite{en-nagre-2024} , Hydrological Drought \cite{jehanzaib-2021} are also assumed by implementing ML. This \cite{belayneh-2013} study shows that these sorts of studies have been frequently conducted based on certain locations. Clustering algorithms, such as K-Means and Gaussian Mixture Models have been introduced in environmental studies around the world. In the example, Authors in \cite{xu-2022} presented the application of the K-Means clustering model for Drought analysis and  authors of \cite{maurer-2021} explored the Gaussian Mixture Models for environmental analysis. These studies highlight the efficacy of clustering techniques in environmental analysis, yet their application to analyze drought conditions in Bangladesh remains underexplored. \\ 
This study tries to address the limitations of previous approaches, particularly their reliance on limited datasets and lack of incorporated critical environmental parameters such as temperature, soil moisture, and humidity, which are vital for a more comprehensive understanding of drought conditions.

\section{Methodologies}
\label{section:methodologies}
This research aims to represent an in-depth analysis of drought conditions across BD, emerging with advanced Machine Learning algorithms to cluster drought characteristics, identify and predict drought conditions. This study leverages a satellite-extracted dataset from \cite{nasa} on the time span of 2012 to 2024 (daily), composed of several environmental factors, i.e. solar radiation, humidity,  temperature, soil moisture and wind-speed. By applying unsupervised machine learning algorithms like K-means clustering, Bayesian Gaussian Mixture, this study categorizes drought severity and tries to provide a framework to understand drought patterns. The objective of this study is to introduce a new classification framework which describes drought severity and predicts the drought scenarios based on satellite based data in Bangladesh.

\subsection{Data Collection}\label{AA}
\begin{figure}[htbp]
	\vspace{0pt}
	\centerline{\includegraphics[height=0.25 \textheight]{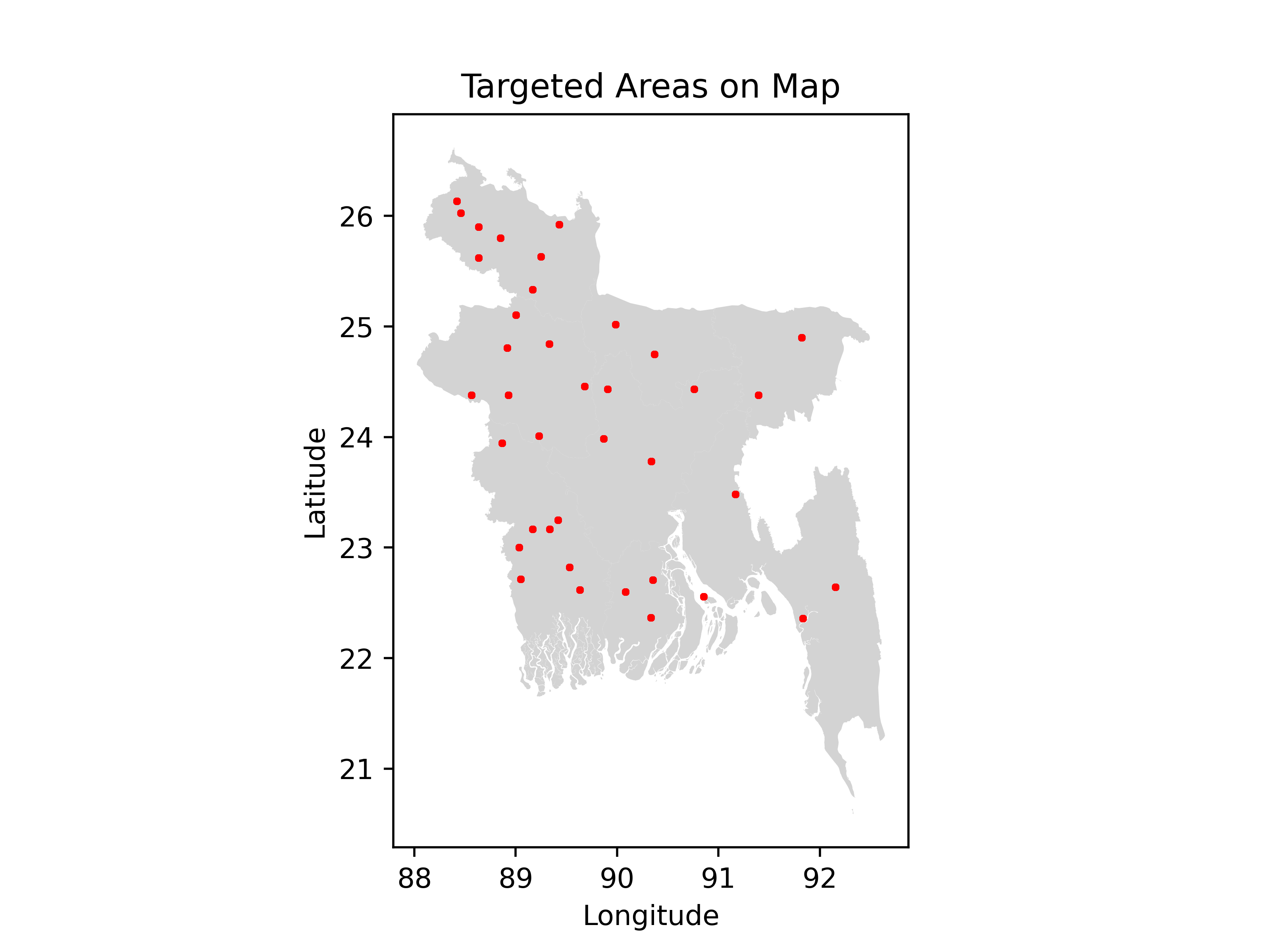}}
	\caption{Data collected across 38 districts of Bangladesh.}
	\label{fig:data_collected_map}
	\label{fig}
\end{figure}

The dataset has been collected from \cite{nasa} across 38 different district-wise locations of Bangladesh based on important weather parameters. The Fig. \ref{fig:data_collected_map} shows the locations that were collected , table \ref{tab:parameters} describes which weather parameters has been introduced to further analysis.

\begin{table}[htbp]
	\centering
	\caption{Model parameters}
	\label{tab:parameters}
	\begin{tblr}{
			hline{1-2,16} = {-}{},
		}
		\textbf{Parameters}                                                 & \textbf{Acronyms}            \\
		Locations                                                           & Latitude, Longitude           \\
		Year (2012-2024)                                                    & Year                          \\
		Day of Year                                                         & DOY                           \\
		{All Sky Surface Shortwave \\Downward \\Irradiance (MJ/m$^2$)/day)} & \textit{ALLSKY\_SFC\_SW\_DWN} \\
		Temperature at 2 Meters (°C)                                        & \textit{T2M}                  \\
		Dew/Frost Point at 2 Meters (°C)                                    & \textit{T2MDEW}               \\
		Earth Skin Temperature (°C)                                         & \textit{TS}                   \\
		Specific Humidity at 2 Meters (g/kg)                                & \textit{QV2M}                 \\
		Relative Humidity at 2 Meters (\%)                                  & \textit{RH2M}                 \\
		Surface Pressure (kPa)                                              & \textit{PS}                   \\
		Wind Speed at 2 Meters (m/s)                                        & \textit{WS2M}                 \\
		Surface Soil Wetness (1)                                            & \textit{GWETTOP}              \\
		Root Zone Soil Wetness (1)                                          & \textit{GWETROOT}             \\
		Profile Soil Moisture (1)                                           & \textit{GWETPROF}             
	\end{tblr}
\end{table}

\begin{figure}[htbp]
	\vspace{-2pt}
	\centering
	\includegraphics[width=0.45\textwidth]{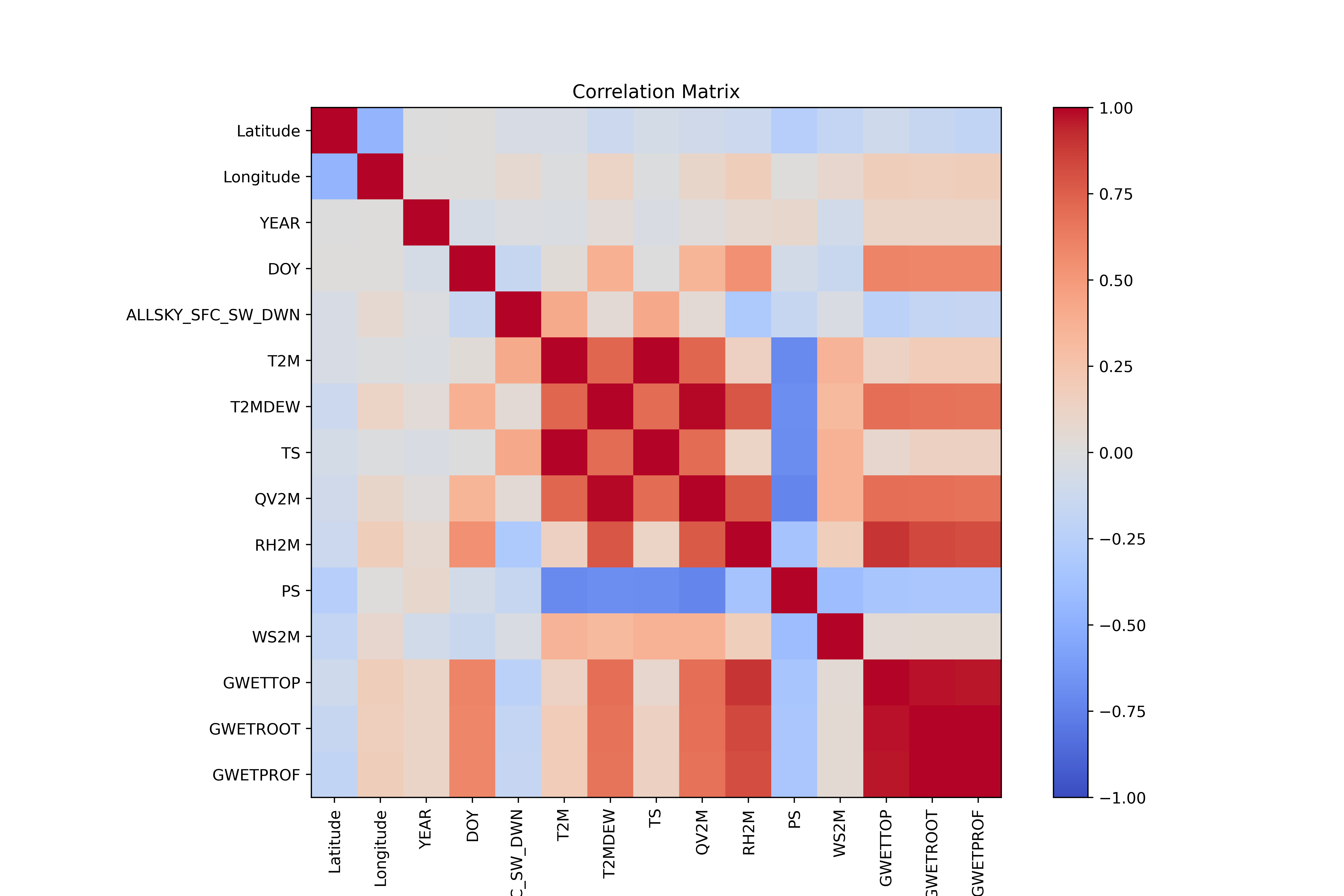}
	\caption{Correlation matrix heatmap of parameters.}
	\label{fig:parameter_heatmap}
\end{figure}

The Fig. \ref{fig:parameter_heatmap} describes the correlation among the model parameters explained in table \ref{tab:parameters}. This figure is a visual representation of their correlation matrix where each cell in the matrix represents the correlation between two different variables while colors indicates the strength and direction of the correlation. Aggregately, this matrix provides valueable insights into the model parameters relationship with each other, which can be helpful for understanding the factors influencing further drought analysis. 

\subsection{Data Pre-Processing}
The raw dataset mentioned in section A was not suitable for Machine Learning(ML) analysis. Before feeding into ML models, this dataset underwent some technical tweaking. 
Sorting and Indexing: The total number of dataset was 38 as because our target district was 38; and after all these data merged into single one. Though the satellite data has some unnecessary labels which might cause a negative impact in ML model, therefore these were also abandoned. 
Unnecessary, NaN(Not a Number) and Null values were abandoned as because the dataset was too heavy, consisting approx 0.17Million rows. The whole dataset was scaled using StandardScaler (uses Standard Deviation to scale). \\
While the entire dataset does not directly describes the drought characteristics; after scaling, the dataset subjected into Unsupervised ML analysis. Then the entire dataset were split into training and testing datasets with the ratio of training and test data being 80:20 in percentages.
\section{Machine Learning Algorithms}
\label{section:ml_algorithms}
\subsection{Clustering}
\label{section:clustering}

\subsubsection{K-Means Clustering}
An unsupervised machine learning algorithm used in clustering analysis to distinguish a particular dataset into non-overlapping ‘k’ distinct. K-means aims to minimize the Within-Cluster-Sum of Squares(WCSS), which is also known as Inertia (\textit{I}).   
\begin{equation}
	I = \sum_{k=1}^{K} \sum_{i=1}^{n_k} \| x_i^{(k)} - \mu_k \|^2
\end{equation}
Where:
- \( K \) is the number of clusters,
- \( n_k \) is the number of points in cluster \( k \),
- \( x_i^{(k)} \) is a data point belonging to cluster \( k \),
- \( \mu_k \) is the centroid of cluster \( k \),
- \( \| x_i^{(k)} - \mu_k \|^2 \) is the squared Euclidean distance between the data point and the centroids.
\begin{equation}
	c_i = \arg\min_k \| x_i - \mu_k \|^2
\end{equation}

\begin{equation}
	\mu_k = \frac{1}{n_k} \sum_{i=1}^{n_k} x_i^{(k)}    
\end{equation}
As shown in Fig.\ref{fig:the_elbow_method} shows the 'Elbow Method' which has been used to minimize the WCSS.

\subsubsection{Bayesian Gaussian Mixture}
Bayesian Gaussian Mixture Model (BGMM) extends the Gaussian Mixture Model(GMM) through a prior distribution. BGMM uses the Dirichlet Process(DP) as a prior, which helps to automatize the number of clusters.  
Unlike other traditional clustering methods, i.e. K-means, it provides a probabilistic assessment. 
\begin{equation}
	p(x) = \sum_{k=1}^{K} \pi_k \mathcal{N}(x|\mu_k, \Sigma_k)
\end{equation}
Where p(x) is the probability density function of GMM, which is a weighted sum of K Gaussian Distribution.
\( \pi_k \) is the mixing coefficient for component \( k \) (with \( \sum_{k=1}^{K} \pi_k = 1 \)). \( \mathcal{N}(x|\mu_k, \Sigma_k) \) is the multivariate Gaussian distribution with mean \( \mu_k \) and covariance matrix \( \Sigma_k \).\\
In BGMM, the Dirichlet Process is replaced with the parameters \( \pi_k \), \( \mu_k \), and \( \Sigma_k \) as:  
\begin{equation}
	\pi \sim \text{Dir}(\alpha/K, \dots, \alpha/K)
\end{equation}
where concentration parameter \( \alpha \) controls the number of clusters. 
However, the BGMMs are estimated on log-likelihood of the data, like Evidence Lower Bound (ELBO): 
\begin{equation}
	\text{ELBO} = \mathbb{E}_q[\log p(X, Z, \theta)] - \mathbb{E}_q[\log q(Z, \theta)]    
\end{equation}
Where \( p(X, Z, \theta) \) and \( q(Z, \theta) \) methodically are the joint probability of the data \( X \), latent variables \( Z \), model parameters \( \theta \) and the approximating the posterior by the variational distribution.

\subsection{Classification}

\subsubsection{K-Nearest Neighbors (KNN)}
A very simple and mainly non-parametric algorithm that is usually used for regression and classification models. This algorithm is based on ‘Euclidean Distance’  
\begin{equation}
	d(p, q) = \sqrt{\sum_{i=1}^{n} (p_i - q_i)^2}
\end{equation}
Where \( p, q \) are two points in \( n \)-dimensional space and \( p_i, q_i \) are the feature values of the two points.
\subsubsection{Naive Bayes}
A probabilistic classifying algorithm based on the ‘Bayes Theorem’  :
\begin{equation}
	P(C_k | x) = \frac{P(x | C_k) P(C_k)}{P(x)}
\end{equation}
Where, \( P(C_k | x) \) is the posterior probability of class \( C_k \) given predictor \( x \),\( P(x | C_k) \) is the likelihood of predictor \( x \) given class \( C_k \), \( P(C_k) \) is the prior probability of class \( C_k \) and \( P(x) \) is the prior probability of predictor \( x \).
On the other hand, Gaussian Naive Bayes uses Gaussian or Normal Distribution  -
\begin{equation}
	P(x_i | C_k) = \frac{1}{\sqrt{2 \pi \sigma_k^2}} \exp\left(-\frac{(x_i - \mu_k)^2}{2 \sigma_k^2}\right) \
\end{equation}
\( \mu_k \) and \( \sigma_k^2 \) are the mean and variance of the feature \( x_i \) in class \( C_k \). 
\subsubsection{Decision Tree}
Decision Tree is an algorithm that uses the 'Gini Impurity' theorem to split the data based on features and creates a tree-like structure.   

\begin{equation}
	Gini(D) = 1 - \sum_{i=1}^{C} p_i^2
\end{equation}

Here, \( C \) is the number of classes and \( p_i \) is the proportion of examples in class \( i \) in dataset \( D \).

\subsubsection{Random Forest}
An ensemble method that builds multiple Decision Trees and merges them together is called Random forest.  \\
After training through machine learning algorithm we have used Kernel Density Estimation (KDE) method to distinguish among clusters for temporal and geospatial analysis.
\section{Result Discussion and Analysis}
\label{section:result_discussion}
\subsection{Cluster Validation}
The dataset underwent through unsupervised Machine Learning’s Clustering algorithms mentioned in \ref{section:clustering} and the Table \ref{tab:silhouette_scores}  represents silhouette scores of them. Silhouette scores offer valuable insights into the clustering results; evaluates the performance metrics of clustering models. K-means score: 0.833 and Bayesian Gaussian Mixture score: 0.749, which validates the clustering approach.

\begin{table}[htbp]
	\centering
	\caption{Silhouette scores of different clustering models}
	\label{tab:silhouette_scores}
	\begin{tblr}{
			column{2} = {c},
			hline{1-2,4} = {-}{},
		}
		\textbf{Model Name}       & \textbf{Silhouette Score (-1 to 1)} \\
		K-Means Clustering        & 0.833                               \\
		Bayesian Gaussian Mixture & 0.749                               
	\end{tblr}
\end{table}

Inspite of their score is almost close, the K-Means model outperforms BGM very well and achieved the silhouette score closer to 1. The K-Means clustering model was accepted for further drought classification analysis.

\begin{figure}[htbp]
	
	\centering
	\includegraphics[height=0.25\textheight]{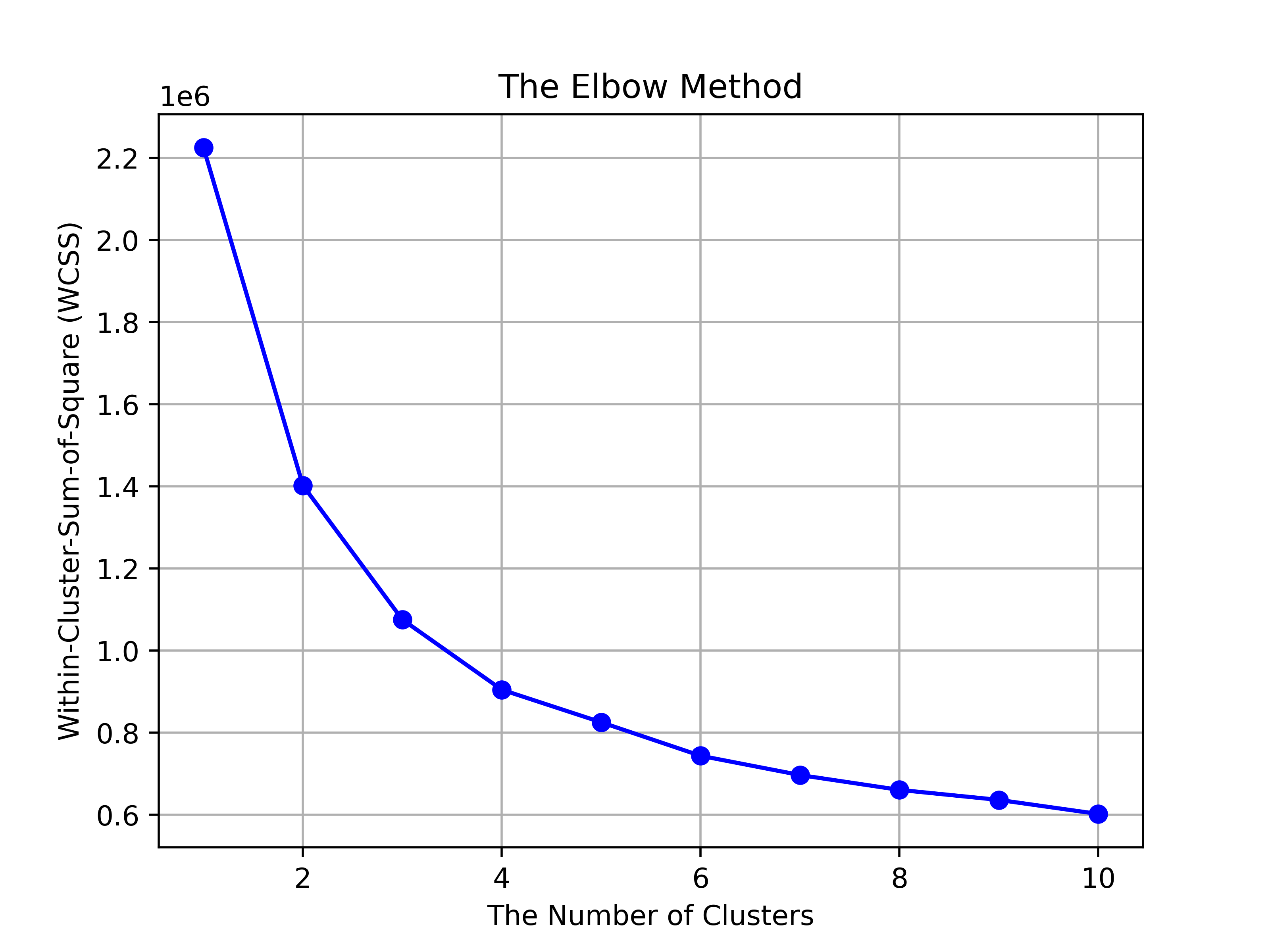}
	\caption{The elbow method.}
	\label{fig:the_elbow_method}
\end{figure}
The elbow method graph from Fig. \ref{fig:the_elbow_method} shows the whole dataset might be able to distinguish three(3) indifferent clusters and the K-Means Clustering algorithm was successful in achieving that with a significant silhouette score. 

\subsection{Cluster Analysis and Interpretation}
\setlength{\intextsep}{0pt}

\begin{figure}[htbp]
	\centering
	\includegraphics[height=0.25\textheight]{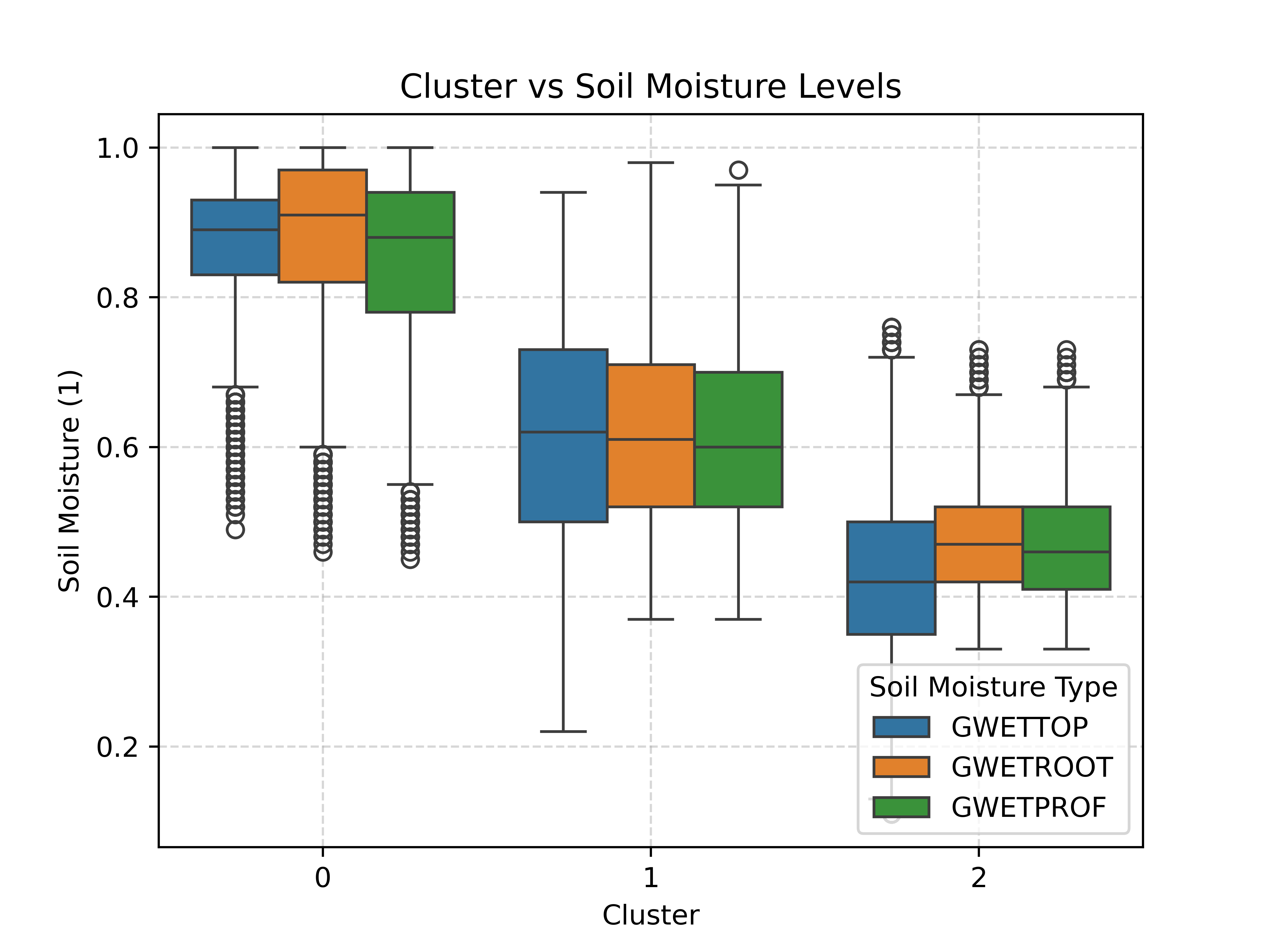}
	\caption{Cluster distribution vs soil moisture boxplot}
	\label{fig: cluster_and_soil_moisture_box_plot}
\end{figure}

\begin{figure}[htbp]
	%\vspace*{-0.5cm}
	\centering
	\includegraphics[height=0.3\textheight]{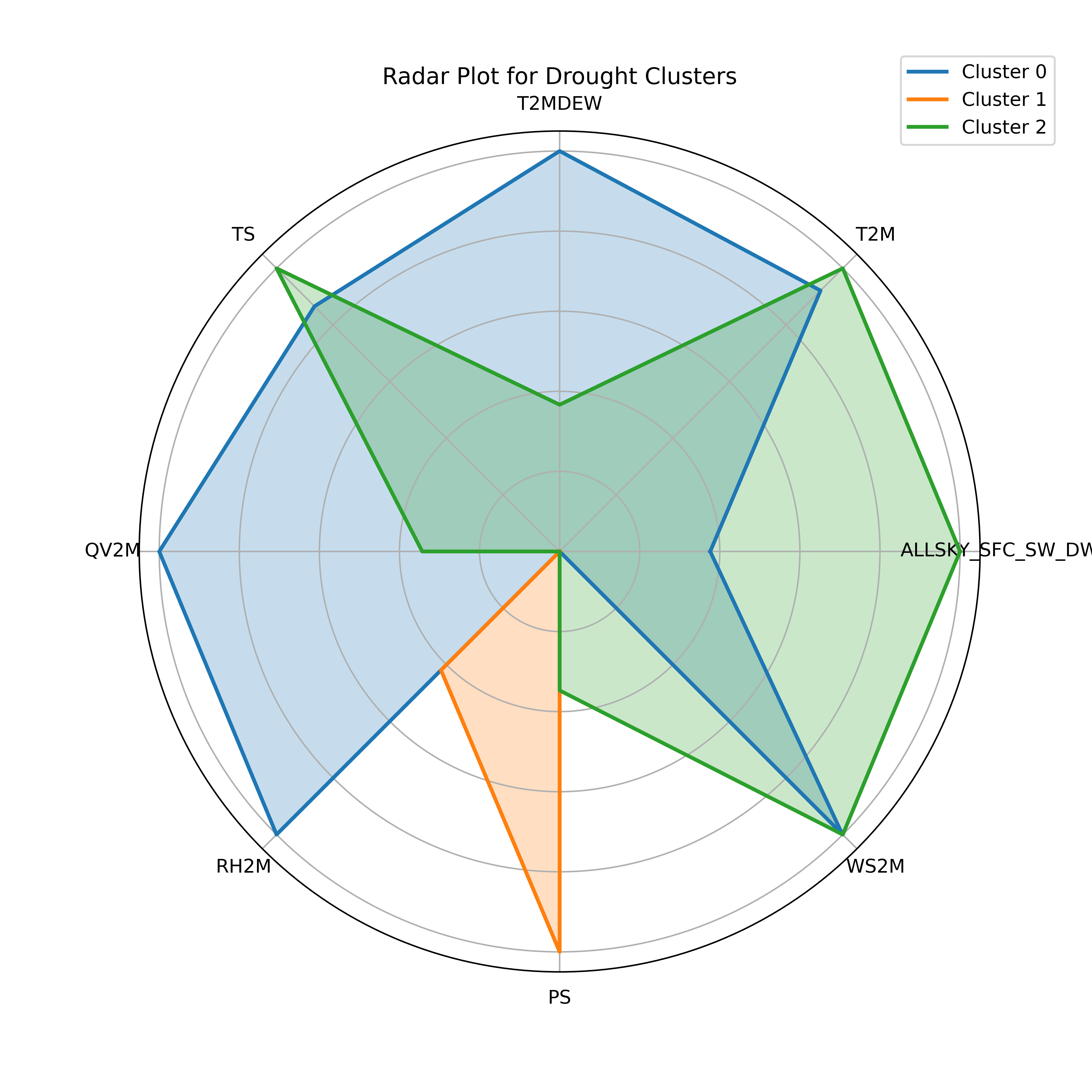}
	\caption{Radar plot of parameters and clusters.}
	\label{fig: radar_plot_combined}
\end{figure}

\begin{figure}[htbp]
	\centering
	\includegraphics[height=0.25\textheight]{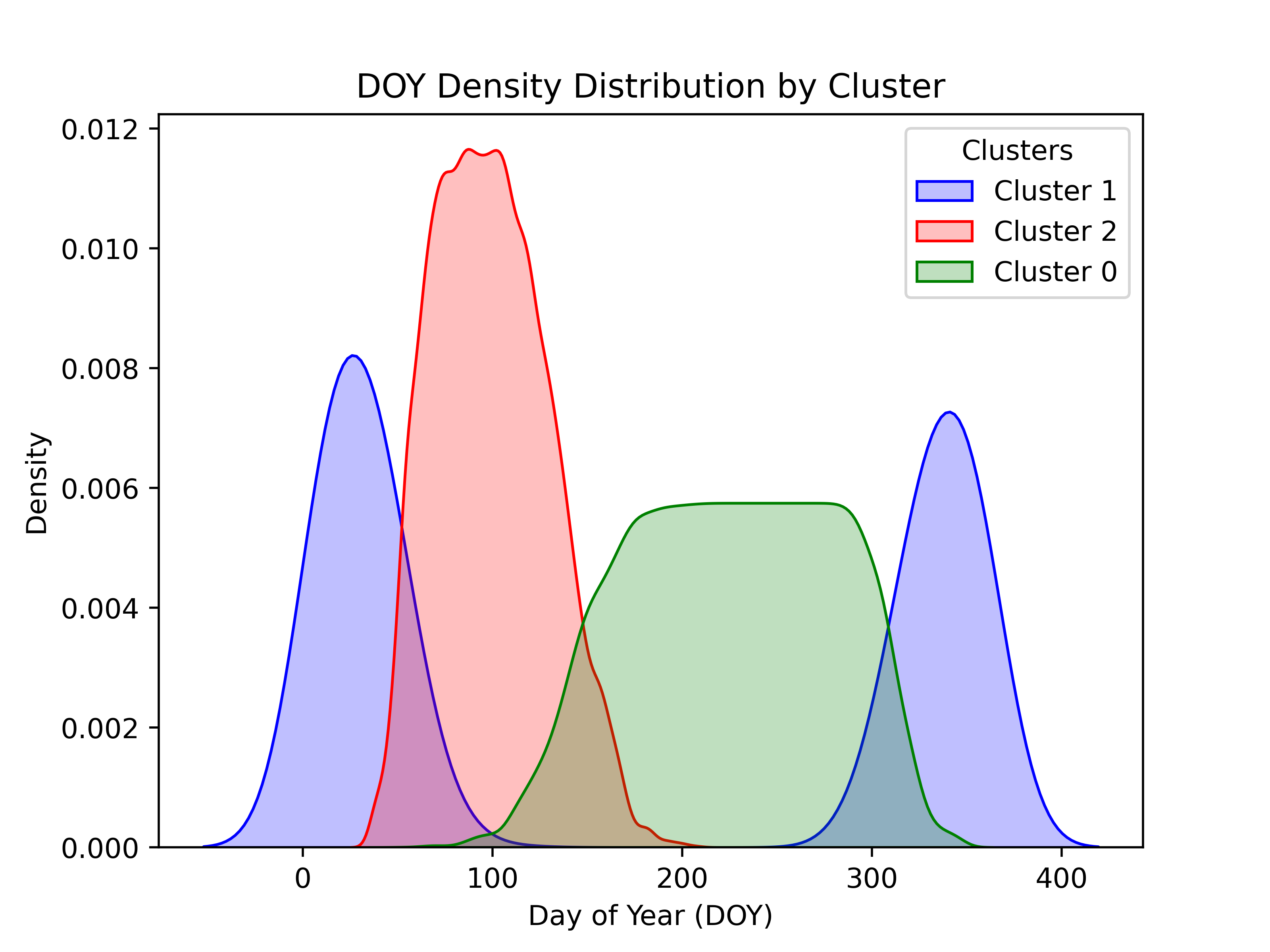}
	\caption{Day wise cluster density}
	\label{fig: day_wise_cluster_density}
\end{figure}

\begin{table*}
	\centering
	\caption{Summary of Cluster Characteristics}
	\label{tab:cluster_summary}
	\begin{tblr}{
			column{odd} = {c},
			column{2} = {c},
			column{4} = {c},
			hlines = {0.05em},
			hline{1,5} = {-}{0.08em},
		}
		\textbf{Cluster}   & \textbf{Extremity} & \textbf{Season}             & {\textbf{Temporal }\\\textbf{Distribution (Days)}} & {\textbf{Soil Moisture }\\\textbf{Levels (Median)}} & \textbf{Key Environmental Characteristics}                                                                                                                                                                                           \\
		\textbf{Cluster 0} & Lower              & Monsoon                     & 150-250                                            & 0.8-0.9                                             & {- Moderate Temperature (T2M) and Dew Point (T2MDEW) \\ - High Relative Humidity (RH2M) and Specific Humidity (QV2M) \\ - High Wind Speed (WS2M) \\ - Low Shortwave Radiation (ALLSKY\_SFC\_SW\_DWN) \\ - Low Surface Pressure (PS)} \\
		\textbf{Cluster 1} & Higher             & Winter                      & 0-50, 250-365                                      & 0.6-0.7                                             & {- Low Temperature (T2M) and Dew Point (T2MDEW) \\ - Very Low Wind Speed (WS2M) \\ - Low Relative Humidity (RH2M) \\ - High Surface Pressure (PS) \\ - Low Shortwave Radiation (ALLSKY\_SFC\_SW\_DWN)}                               \\
		\textbf{Cluster 2} & Moderate           & {Transitional/\\Dry Season} & 50-200                                             & 0.4-0.5                                             & {- Moderate to Low Temperature (T2M) and Dew Point (T2MDEW) \\ - Lowest Relative Humidity (RH2M) \\ - High Shortwave Radiation (ALLSKY\_SFC\_SW\_DWN) \\ - High Wind Speed (WS2M) \\ - High Surface Pressure (PS)}                   
	\end{tblr}
\end{table*}

\begin{figure*}[htbp]  % Forceful placement in two-column layout
	\vspace{0pt}
	\centering
	\includegraphics[height=0.2\textheight]{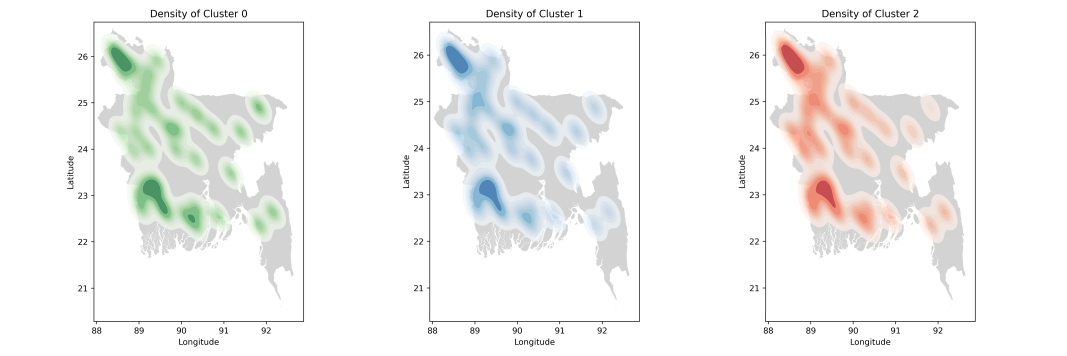}
	\caption{Cluster densities in different locations}
	\label{fig:clusters_density_location}
\end{figure*}

Fig. \ref{fig: cluster_and_soil_moisture_box_plot} represents the relationship between clusters and soil moisture parameters. Fig. \ref{fig: radar_plot_combined} illustrates the relationship of the overall parameters with the clusters, and Fig. \ref{fig: day_wise_cluster_density} shows the day-wise temporal distribution of clusters  over the year. In Table \ref{tab:cluster_summary}, we have explained our detailed analysis based on these graphs and the interpretation of clusters. Now Based on the Temporal Distribution, Soil Moisture Levels and other Environmental Characteristics which has been described in the table \ref{tab:cluster_summary}, as well as we have categorized the clusters into three different extrimity sections and mentioned there. 

\subsection{Geospatial observations of Clusters}

Fig.\ref{fig:clusters_density_location} describes the density distribution of three clusters across Bangladesh. Each cluster is represented by three \\different colors (Green, Blue and Red). Colors are only the representations, not the indicators of their intensiveness. \\
Firstly, all three clusters are showing high densities in the north-west locations in Bangladesh which indicates that these regions experience all three types of drought conditions over the year. Secondly, the clusters are significantly showing lower densities in the east and southeast regions, indicating these areas might experience less vulnerable drought conditions. Thirdly, the central regions are showing variable density across the all three clusters, which indicates they might experience transitional drought conditions. Finally, the cluster distribution seems constantly lower in the south-East coastal regions, addressing these areas might face fewer drought conditions over time. Furthermore, this distribution pattern indicates the patterns are not uniformly distributed, indicating the drought conditions significantly vary across different locations of Bangladesh.

\subsection{Classification Model Validation}

\begin{table}[htbp]
	\centering
	\caption{Confusion matrices and accuracy rates of different classifiers}
	\label{tab:classifiers}
	\begin{tblr}{
			row{odd} = {c},
			row{2} = {c},
			row{4} = {c},
			hline{1,7} = {-}{0.08em},
			hline{2} = {-}{},
		}
		\textbf{Classifiers} & \textbf{Confusion Matrix}                                                                       & \textbf{Accuracy Rate} \\
		Decision Tree        & $ \begin{bmatrix} 14746 & 872 & 263 \\ 153 & 9561 & 400 \\ 151 & 267 & 7825 \end{bmatrix} $     & 91\%                   \\
		Random Forest        & $ \begin{bmatrix} 14819 & 639 & 423 \\ 247 & 9512 & 355 \\ 239 & 465 & 7639 \end{bmatrix} $     & 92\%                   \\
		KNN                  & $ \begin{bmatrix} 13806 & 1344 & 731 \\ 869 & 8221 & 1324 \\ 410 & 1525 & 6708 \end{bmatrix} $  & 84\%                   \\
		Naive Bayes          & $ \begin{bmatrix} 14199 & 1092 & 490 \\ 307 & 9328 & 479 \\ 287 & 709 & 7247 \end{bmatrix} $    & 86\%                   \\
		&                                                                                                 &                        
	\end{tblr}
\end{table}

Table \ref{tab:classifiers} shows the confusion matrix and accuracy score of our 4 different ML classification models based on their drought cluster prediction which we’ve previously interpreted through clustering methodology. Here, each of the ML models seems to be performing very well. From these four models, Random Forest over-performing(92\%) all other models with the highest accuracy and robustness in this scenario. 
The diagonal elements of the confusion matrix show the count of correctly classified instances of each cluster. Random forest has successfully classified 14819,  9512 and 7639 data correctly for Cluster 0, Cluster 1 and Cluster 2. However, the performance of the confusion matrix of Random Forest is also significantly higher than others which shows very few misclassifications across all the three classes ensuring an overall balanced performance. 
Decision Tree can also be considered as an acceptable model along with Random Forest, performing 91\% of accuracy and might offer good interpretability. 
On the other hand, simpler ML models like KNN and Naive Bayes compared with complex models like Random Forest and Decision Tree; their underperformance showing the complexity of the dataset for understanding the drought patterns.

\section{Conclusion}
\label{section:conclusion}

In this research, we developed an efficient approach for classifying drought intensity using unsupervised machine learning algorithms. By employing K-Means and Bayesian Gaussian Mixture algorithms, we effectively classified drought into three distinct levels: high, moderate, and low -  across 38 districts in Bangladesh. We have also determined the seasonal appearance of these drought clusters accordingly. This analysis highlighted a significant spatial variability in drought vulnerabilities, where the northwestern regions being prone to severe drought vulnerabilities with the eastern and south-eastern districts remain less affected. 
Furthermore, this study demonstrates the effectiveness of unsupervised learning in predicting and classifying drought using satellite data. Looking ahead, future research could focus on refining the model with additional environmental parameters and exploring its applicability in other climate-sensitive regions. The use of machine learning in drought analysis represents a promising avenue for tackling one of Bangladesh's most pressing environmental challenges, contributing to more resilient agricultural and water management systems. This framework can significantly assist in addressing the hazards associated with drought in Bangladesh incorporating this model into national and regional policy frameworks for water resource management and agricultural planning.

\end{document}